\documentclass[10pt,twocolumn]{article}

\usepackage[margin=0.72in,columnsep=0.25in]{geometry}
\usepackage[T1]{fontenc}
\usepackage{lmodern}
\usepackage{microtype}
\usepackage{amsmath,amssymb}
\usepackage{booktabs}
\usepackage{graphicx}
\usepackage{multirow}
\usepackage{array}
\usepackage{enumitem}
\usepackage{natbib}
\usepackage{url}
\usepackage[hidelinks,hypertexnames=false]{hyperref}
\usepackage{xcolor}
\usepackage{listings}
\usepackage{algorithm}
\usepackage{algpseudocode}
\usepackage{placeins}

\setlist{nosep,leftmargin=*}
\newcommand{\model}[1]{\texttt{#1}}
\newcommand{\pct}[1]{#1\%}
\newcommand{\dataset}{\textsc{SLM-ToolCall-100}}

\lstdefinelanguage{json}{
  basicstyle=\ttfamily\footnotesize,
  showstringspaces=false,
  breaklines=true,
  frame=single,
  literate={\{}{{{\color{black}\{}}}{1} {\}}{{{\color{black}\}}}}{1}
}

\title{Beyond Fluent Generation: A CPU Reliability Benchmark for \\ MCP-Style Tool Calling in Sub-2B Small Language Models for Edge Deployment}
\author{Abrar Shahriar \qquad Qurat-Ul-Ain Mastoi \\[-0.15em]
\small University of the West of England, Bristol,BS16 1QY United Kingdom\\
\small \texttt{abrar2.shahriar@live.uwe.ac.uk}\\
\small \texttt{Qurat-Ul-Ain.Mastoi@uwe.ac.uk}\\}
\date{}

\begin{document}
\maketitle

\begin{abstract}
Resource-constrained edge computers including Raspberry Pi, NVIDIA Jetson Nano, Arduino UNO Q, Orange Pi, and LattePanda platforms motivate local small language model (SLM) agents that can operate with reduced cloud dependence, improved data locality, and intermittent connectivity. Yet Model Context Protocol (MCP)-style tool use requires more than fluent generation: an agent must emit machine-readable structure, identify the intended tool, reproduce correct arguments, and avoid unintended actions. This study establishes a platform-agnostic CPU baseline by evaluating five open-weight models below two billion parameters Phi-1.5, Pythia-1.4B, TinyLlama-1.1B-Chat, Qwen2.5-0.5B, and Qwen2.5-1.5B on 100 controlled prompts spanning weather retrieval, web search, calculation, email composition, and task creation. Each checkpoint is tested with greedy decoding and nucleus sampling. The recovery parser removes Markdown fences, extracts a brace-delimited substring, and measures parseability, tool-name correctness, argument presence, and value agreement. Under this recovered-call criterion, Qwen2.5-1.5B reaches \pct{75} with greedy decoding and \pct{79} with sampling, while Qwen2.5-0.5B reaches \pct{72} and \pct{32}, respectively. Phi-1.5 scores \pct{0} in both conditions, and Pythia and TinyLlama remain at or below \pct{7}. A strict post-hoc audit finds that only 5 of 1,000 complete raw responses are directly parseable as JSON, demonstrating strong dependence on output recovery. In a separate three-prompt CPU probe, Qwen2.5-1.5B records the highest process memory (7,960 MiB) and mean latency (30.782 s), whereas Qwen2.5-0.5B uses 3,637 MiB and averages 10.627 s. The results expose a reliability resource trade-off relevant to prospective edge deployment, but they do not constitute direct measurements on the named boards or a complete MCP implementation. Safe deployment requires schema validation, constrained generation or repair, least-privilege execution, and escalation for consequential actions.
\end{abstract}

\noindent\textbf{Keywords:} small language models; tool calling; structured generation; edge inference; JSON; agent reliability

\section{Introduction}

Language models can extend their capabilities by invoking calculators, search systems, databases, and other external services. Early tool-use work demonstrated that a model can learn when and how to call an API \citep{schick2023toolformer}, while API-focused systems showed that correct tool names and arguments remain difficult even for substantially larger models \citep{patil2023gorilla}. Once an output can trigger an external action, fluent text is no longer sufficient: a malformed call can halt a workflow, and a syntactically valid but semantically wrong call can modify state, disclose information, or contact the wrong recipient.

SLMs offer a compelling deployment alternative for edge computing devices. Local inference may reduce network dependence and data exposure, but it remains constrained by memory capacity, bandwidth, compute, energy, and thermal limits. Surveys and edge benchmarks consequently evaluate both capability and systems cost rather than treating parameter count as a complete proxy for deployability \citep{lu2025slm,chen2025elib}. Application-specific training may also allow a narrow model to outperform a larger general model on a bounded task \citep{kandala2024tinyllm,jhandi2025toolcalling}. These observations motivate a joint question aligned with the title of this study: how reliably can an unmodified sub-2B model produce MCP-style tool calls, and what CPU resource cost accompanies that reliability?

This study makes four contributions. First, it presents a controlled benchmark of JSON-based tool-call generation for five open-weight models below 2B parameters. Second, it separates recoverable syntax, tool identification, argument completion, value agreement, and extra-argument control. Third, it compares greedy and sampling decoding on the same 100 prompts and reports paired tests as well as Wilson confidence intervals. Fourth, it reports a separate local CPU resource probe and explicitly audits the gap between whole-response strict JSON and JSON recovered by an extraction layer.

The experiment is best understood as a controlled, MCP-inspired function-call baseline, not as a full Model Context Protocol (MCP) evaluation. It does not perform tool discovery, client/server messaging, live execution, or multi-step trajectories. This distinction is important because MCP benchmarks include long tool descriptions, variable execution outcomes, intermediate observations, and both single- and multi-step calls \citep{fan2025mcptoolbench}.

The prospective hardware landscape is heterogeneous. Raspberry Pi 5 and Orange Pi 5 are Arm-based single-board computers, while NVIDIA Jetson Nano is an embedded platform designed for AI applications \citep{raspberrypi2026,orangepi2026,nvidia2026jetson}. Arduino UNO Q combines a Debian-capable microprocessor environment with a real-time microcontroller, and LattePanda provides x86 single-board computer families \citep{arduino2026unoq,lattepanda2026}. These differences in architecture, acceleration, memory, operating environment, and power constraints make it unsafe to infer board-level deployability from parameter count alone. The present CPU experiment therefore supplies a controlled reliability and resource baseline for future device-specific evaluation; Raspberry Pi, NVIDIA Jetson Nano, Arduino UNO Q, Orange Pi, and LattePanda were not directly tested.

\section{Related Work}

\subsection{Small models and edge inference}

\citet{lu2025slm} survey decoder-only SLMs from 100M to 5B parameters and benchmark latency and memory, emphasizing that architecture, context, software, quantisation, and hardware jointly determine runtime cost. \citet{kandala2024tinyllm} similarly argue for application-specific small models in edge sensing settings, where remote inference can introduce latency, connectivity, and privacy costs. ELIB broadens edge evaluation to throughput, latency, accuracy, floating-point operations, and memory-bandwidth utilisation across heterogeneous platforms \citep{chen2025elib}. Our study follows this systems-aware perspective but focuses on structured action generation rather than general language quality.

The evaluated model families were created for different purposes. Phi-1.5 uses synthetic ``textbook-quality'' data to improve reasoning at approximately 1.3B parameters \citep{li2023phi}. Pythia provides controlled training checkpoints intended for analysis across scales \citep{biderman2023pythia}. TinyLlama is a 1.1B model trained on roughly one trillion tokens and shares architectural choices with Llama 2 \citep{zhang2024tinyllama}. Qwen2.5 reports improvements in instruction following and structured-data handling across a family of sizes \citep{qwen2024qwen25}. Because the present checkpoint set mixes base and chat-oriented models, the comparison measures deployable checkpoint behaviour, not a clean causal effect of parameter count.

Peer-reviewed on-device studies provide important comparators. MobileLLM shows at ICML 2024 that architecture design, rather than parameter count alone, materially affects sub-billion-parameter quality and reports competitive API-calling behaviour for mobile-oriented models \citep{liu2024mobilellm}. TinyAgent, presented at EMNLP 2024, combines task-specific function-calling fine-tuning, tool retrieval, and quantisation for local agent execution \citep{erdogan2024tinyagent}. Octopus extends this direction at NAACL 2025 by training 2B--7B on-device models on 30,000 API calls and using conditional masking to reduce output-format errors \citep{chen2025octopus}. These systems differ from our untuned-checkpoint baseline: their stronger results demonstrate the value of specialised data and output-control mechanisms rather than establishing that arbitrary small checkpoints are inherently reliable.

\subsection{Function-calling evaluation}

Tool-use evaluation has progressed from selecting and formatting single API calls to realistic multi-turn interaction. Hammer introduces function masking and irrelevant-function training for robust on-device function calling, motivated in part by sensitivity to naming conventions and distractor functions \citep{lin2024hammer}. HammerBench evaluates incomplete requests, changing intent, argument edits, indirect references, and fine-grained mobile-assistant interactions; parameter-name errors are a prominent failure mode \citep{wang2025hammerbench}. MCPToolBench++ extends evaluation to large collections of MCP tools and both single- and multi-step calls \citep{fan2025mcptoolbench}. Robustness work further shows that performance can change under natural query variation or when semantically related tools are added \citep{rabinovich2025robustness}.

Published benchmarks also clarify how the present controlled task differs from realistic tool use. API-Bank provides an executable environment with 73 APIs and dialogues requiring planning, retrieval, and invocation \citep{li2023apibank}. ToolSandbox evaluates stateful execution, implicit dependencies, insufficient-information cases, and intermediate trajectory milestones \citep{lu2025toolsandbox}. At ICLR 2025, $\tau$-bench introduced tool agent user interaction in policy-governed domains and showed that successful tool calls must be evaluated within a continuing conversation rather than as isolated strings \citep{yao2025taubench}. Together, these works support retaining our benchmark as a syntax-and-argument diagnostic while avoiding claims of complete agent competence.

These benchmarks expose a limitation of the present task: every prompt states the desired tool and all target values explicitly. The experiment therefore tests instruction adherence, copying, and structured serialization more directly than intent recognition or planning. Its value is as a controlled lower-complexity diagnostic: failure even in this favourable setting is evidence against unguarded deployment.

\subsection{Structured output and safety}

An operational tool call must satisfy multiple layers: the response must be extractable, parseable, schema-conformant, and semantically appropriate. Constrained decoding can guarantee some syntactic properties, but framework coverage, generation quality, and efficiency still require empirical evaluation \citep{geng2025jsonschema}. Safety adds another layer. ToolEmu demonstrates that agent failures can lead to privacy and financial harms and motivates sandboxed evaluation before live execution \citep{ruan2023toolemu}. The present benchmark does not evaluate adversarial safety or harmful requests; it tests whether benign, fully specified calls survive the serialization pipeline.

\section{Method}

\begin{figure*}[p]
  \centering
  \includegraphics[width=0.58\textwidth]{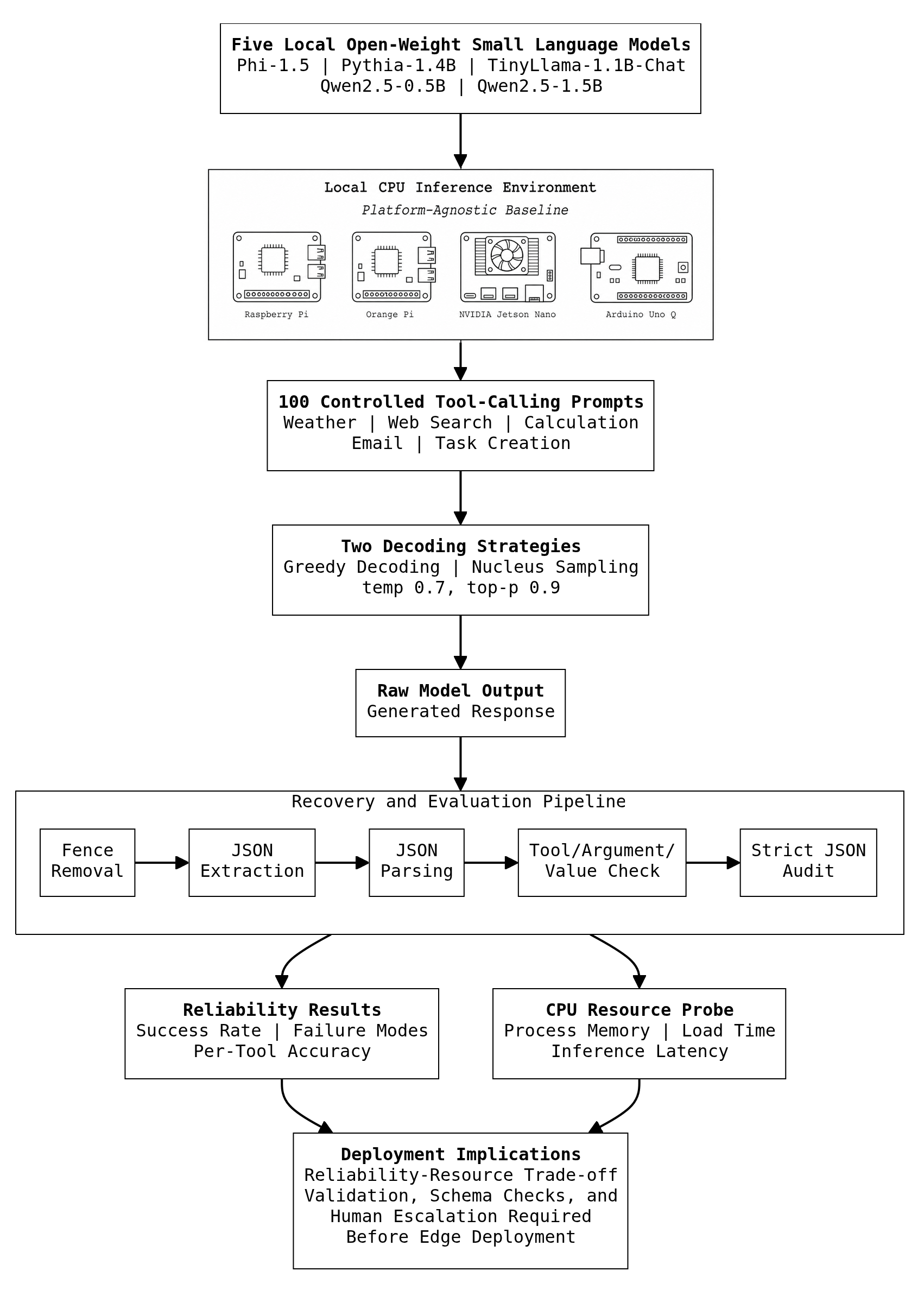}
  \caption{Overview of the local SLM tool-calling benchmark. Raspberry Pi, Orange Pi, NVIDIA Jetson Nano, and Arduino UNO Q are shown as prospective edge deployment targets; LattePanda is an additional target discussed in the text. None of these devices was directly benchmarked in this study.}
  \label{fig:workflow-overview}
\end{figure*}

Figure~\ref{fig:workflow-overview} summarises the complete experimental workflow, from checkpoint selection and controlled prompt generation to decoding, JSON recovery, reliability scoring, resource measurement, and deployment interpretation. The hardware platforms shown in the diagram are prospective targets that motivate the study; the reported experiment used the platform-agnostic CPU procedure described below.

\subsection{Models and experimental design}

Table~\ref{tab:models} lists the exact Hugging Face checkpoint identifiers and parameter counts recorded by the resource notebook. All models were loaded through \texttt{transformers} with remote model code permitted. The main experiment generated 150 new tokens per prompt. Greedy decoding used \texttt{do\_sample=False}; sampling used temperature 0.7 and nucleus probability $p=0.9$. The global random seed was 42. Models ran in floating-point precision (FP32 on CPU); no quantisation was enabled.

\begin{table}[t]
\centering
\caption{Evaluated checkpoints. Parameter counts are those computed in the notebook.}
\label{tab:models}
\small
\begin{tabular}{@{}p{0.70\columnwidth}r@{}}
\toprule
Checkpoint & Parameters (M) \\
\midrule
\model{microsoft/phi-1\_5} & 1,418.27 \\
\model{EleutherAI/pythia-1.4b-deduped} & 1,414.65 \\
\model{TinyLlama/TinyLlama-1.1B-Chat-v1.0} & 1,100.05 \\
\model{Qwen/Qwen2.5-0.5B} & 494.03 \\
\model{Qwen/Qwen2.5-1.5B} & 1,543.71 \\
\bottomrule
\end{tabular}
\end{table}

\subsection{Benchmark prompts}

The \dataset{} dataset contains 100 English prompts, balanced across five mock tools (20 per tool): \texttt{get\_weather}, \texttt{search\_web}, \texttt{calculate}, \texttt{send\_email}, and \texttt{create\_todo}. Each prompt names one tool, enumerates every expected argument and value, and requests only an object with top-level keys \texttt{tool} and \texttt{parameters}. Prompts vary locations, units, queries, result counts, expressions, email fields, task titles, due dates, and priorities. Their order was shuffled using seed 42 and then reused for every model strategy condition.

For example, a prompt requests the calculation tool with an expression, and the target is:
\begin{lstlisting}[language=json]
{"tool":"calculate",
 "parameters":{"expression":"12+18/3"}}
\end{lstlisting}
Tools were simulated and were not executed in the main benchmark. Thus, ``success'' means agreement with a deterministic expected call, not successful completion of an external action.

\subsection{Parsing and evaluation}

The notebook's parser first removes Markdown code-fence markers, then greedily extracts text from the first opening brace to the last closing brace, and finally applies \texttt{json.loads}. We call this \emph{recovered JSON validity}. For a parsed object, evaluation proceeds sequentially:

\begin{enumerate}
  \item the \texttt{tool} value must equal the target name;
  \item \texttt{parameters} must exist and contain all expected keys;
  \item string values must match case-insensitively and numeric values within relative tolerance $10^{-3}$;
  \item extra parameter keys are recorded separately.
\end{enumerate}

The notebook's \texttt{fully\_correct} outcome requires recovered validity, correct tool, all expected parameters, and matching values, but it does \emph{not} require the absence of extra parameters. We therefore report both notebook success and a post-hoc \emph{schema-exact} outcome that additionally requires no extra parameter keys. We also re-parse each entire stripped response without fence removal or substring extraction to measure \emph{strict whole-response JSON validity}. This audit is computed from the saved responses and does not alter model outputs.

For each rate $\hat p=x/n$, we report a two-sided 95\% Wilson interval. Because both strategies use the same prompts, decoding differences are additionally tested with an exact McNemar test on discordant prompt outcomes. These tests are descriptive and are not corrected for multiple comparisons.

\subsection{CPU resource probe}

A separate notebook cell loads each unquantised model on CPU with FP32 weights and records load time and resident set size (RSS) for the complete Python process after loading. Following one warm-up generation, it measures three greedy generations capped at 50 new tokens and reports their mean and population standard deviation. The CPU model, core count, RAM specification, model revisions, and exact software versions were not captured. RSS was measured in a shared, long-running process rather than an isolated fresh process. Resource results are therefore machine- and protocol-specific indicators, not reproducible hardware-normalised benchmarks.

\begin{algorithm*}[t]
\caption{Evidence-producing workflow implemented in the notebook}
\label{alg:workflow}
\begin{algorithmic}[1]
\Require Checkpoints $\mathcal{M}$; tools $\mathcal{T}$; 20 argument sets per tool; strategies $\mathcal{S}=\{\textsc{Greedy},\textsc{Sampling}\}$; seed 42
\Ensure Detailed responses and stage-wise scores; Wilson intervals; error counts; CPU resource measurements
\State Seed Python, NumPy, and PyTorch random-number generators with 42
\ForAll{$t\in\mathcal{T}$}
  \State Shuffle the 20 argument sets for $t$
  \ForAll{argument sets $a$ assigned to $t$}
    \State Construct a prompt that names $t$, supplies $a$, and requests only a JSON object
    \State Store target $y\gets\{\texttt{tool}:t,\texttt{parameters}:a\}$
  \EndFor
\EndFor
\State Combine the five balanced tool subsets into 100 prompt target pairs
\ForAll{checkpoints $m\in\mathcal{M}$}
  \ForAll{strategies $s\in\mathcal{S}$}
    \State Load the tokenizer and causal language model for $m$
    \State Set $s$ to greedy decoding, or sampling with temperature 0.7 and top-$p=0.9$
    \ForAll{prompt target pairs $(x,y)$}
      \State Tokenise $x$ (maximum input length 512) and generate at most 150 new tokens
      \State Decode the response and remove an echoed prompt prefix when present
      \State Remove Markdown fence markers and extract the brace-delimited candidate substring
      \State Attempt JSON parsing; if it fails, record \emph{JSON parse error} and continue
      \State Check the tool name, required argument keys, value agreement, and extra keys in order
      \State Record all Boolean stage indicators, the terminal error label, prompt, target, and response
    \EndFor
    \State Release the model and clear available accelerator cache
  \EndFor
\EndFor
\State Aggregate notebook success and compute a 95\% Wilson interval for each model strategy pair
\State Tabulate terminal errors and compare paired greedy/sampling outcomes with exact McNemar tests
\State Re-parse each complete raw response without recovery to obtain strict whole-response JSON validity
\ForAll{checkpoints $m\in\mathcal{M}$ in the separate CPU probe}
  \State Load $m$ on CPU in FP32; record elapsed load time, process RSS, and parameter count
  \State Run one 30-token warm-up generation
  \State Time three greedy generations capped at 50 new tokens
  \State Store mean and population standard deviation of latency; release model and collect garbage
\EndFor
\State Save detailed results, summary tables, error counts, and visualisations
\end{algorithmic}
\end{algorithm*}

Algorithm~\ref{alg:workflow} covers every procedure that produced the primary evidence reported in this paper. The notebook also contains a later exploratory LangGraph demonstration. Because the benchmark variables were unavailable in that cell, it fell back to five prompts and two models. We report this run separately as an integration diagnostic rather than combining it with the model benchmark.

\section{Results}

\subsection{End-to-end recovered calls}

Table~\ref{tab:mainresults} reports all ten model strategy conditions. Qwen2.5-1.5B sampling is highest at \pct{79} (95\% CI \pct{70.0}--\pct{85.8}), followed by its greedy condition at \pct{75} and Qwen2.5-0.5B greedy at \pct{72}. Sampling sharply reduces Qwen2.5-0.5B success to \pct{32}. The remaining checkpoints score between \pct{0} and \pct{7}.

\begin{table*}[t]
\centering
\caption{Recovered-call outcomes on 100 prompts per condition. CI is the 95\% Wilson interval. ``Exact'' additionally rejects extra parameter keys. Strict JSON parses the complete raw response with no recovery.}
\label{tab:mainresults}
\small
\begin{tabular}{@{}llrrrrr@{}}
\toprule
Model & Strategy & Success & Rate & 95\% CI & Exact & Strict JSON \\
\midrule
Pythia-1.4B & greedy & 0 & \pct{0} & [\pct{0.0}, \pct{3.7}] & \pct{0} & \pct{0} \\
Pythia-1.4B & sampling & 3 & \pct{3} & [\pct{1.0}, \pct{8.5}] & \pct{2} & \pct{0} \\
Qwen2.5-0.5B & greedy & 72 & \pct{72} & [\pct{62.5}, \pct{79.9}] & \pct{72} & \pct{0} \\
Qwen2.5-0.5B & sampling & 32 & \pct{32} & [\pct{23.7}, \pct{41.7}] & \pct{31} & \pct{0} \\
Qwen2.5-1.5B & greedy & 75 & \pct{75} & [\pct{65.7}, \pct{82.5}] & \pct{75} & \pct{0} \\
Qwen2.5-1.5B & sampling & 79 & \pct{79} & [\pct{70.0}, \pct{85.8}] & \pct{79} & \pct{5} \\
TinyLlama-1.1B-Chat & greedy & 0 & \pct{0} & [\pct{0.0}, \pct{3.7}] & \pct{0} & \pct{0} \\
TinyLlama-1.1B-Chat & sampling & 7 & \pct{7} & [\pct{3.4}, \pct{13.7}] & \pct{7} & \pct{0} \\
Phi-1.5 & greedy & 0 & \pct{0} & [\pct{0.0}, \pct{3.7}] & \pct{0} & \pct{0} \\
Phi-1.5 & sampling & 0 & \pct{0} & [\pct{0.0}, \pct{3.7}] & \pct{0} & \pct{0} \\
\bottomrule
\end{tabular}
\end{table*}

\begin{figure*}[t]
  \centering
  \includegraphics[width=0.93\textwidth]{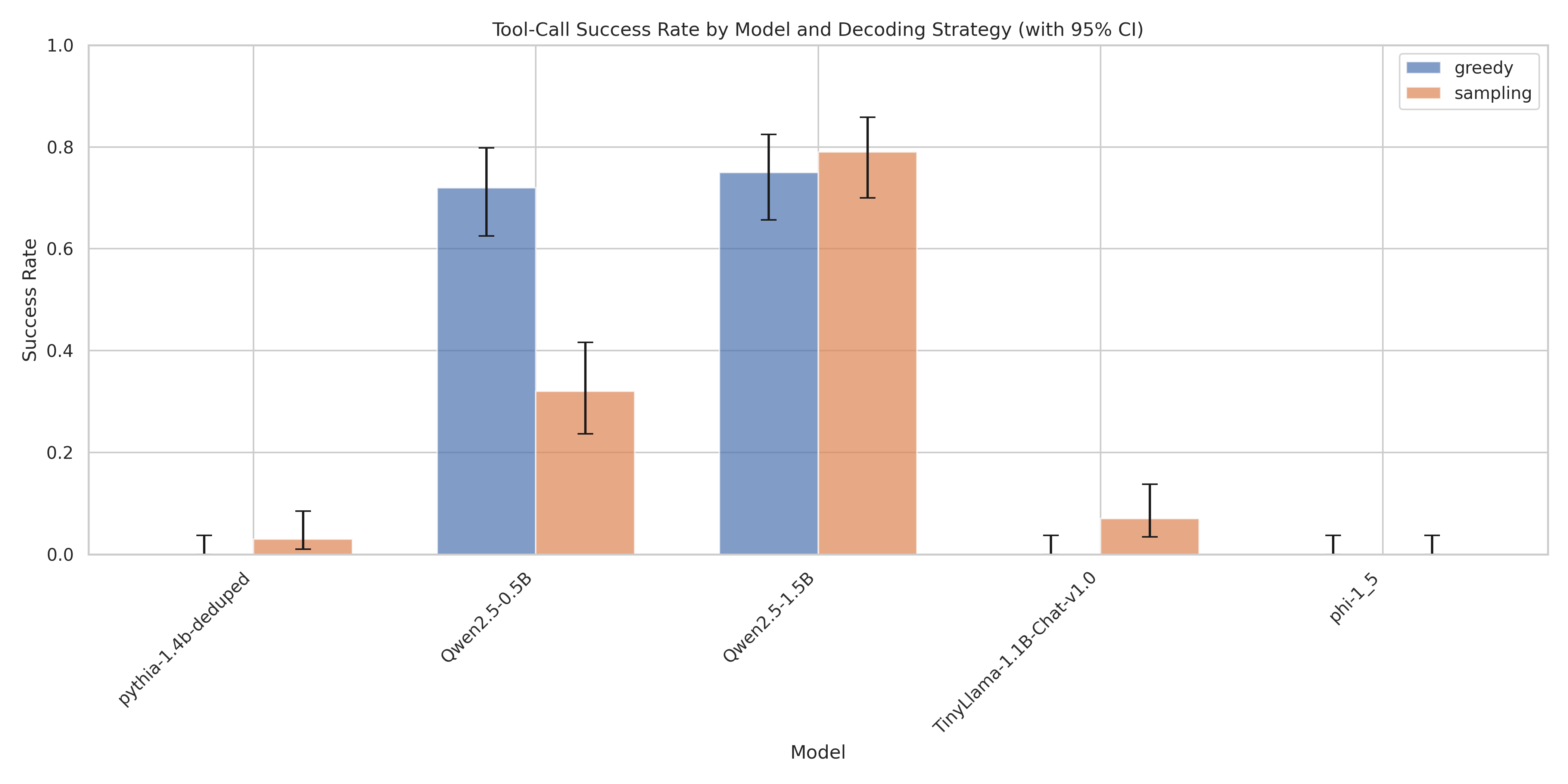}
  \caption{Notebook success rates with 95\% Wilson confidence intervals. Success is based on recovered rather than whole-response JSON.}
  \label{fig:success}
\end{figure*}

The paired comparison confirms that the decoding effect is model-dependent. For Qwen2.5-0.5B, 50 prompts succeed only under greedy decoding and 10 only under sampling (exact McNemar $p=1.62\times10^{-7}$). For Qwen2.5-1.5B, the discordant counts are 15 and 19 ($p=.608$), providing no evidence of a difference in this sample. TinyLlama has seven sampling-only successes ($p=.0156$); Pythia has three ($p=.25$); Phi has no discordant successes. Since sampling was run only once, these outcomes do not estimate run-to-run stochastic variance.

\subsection{Component metrics and strict-output audit}

Table~\ref{tab:components} shows where the sequential pipeline fails. Qwen2.5-0.5B greedy produces recoverable JSON in \pct{82} of cases, falling to \pct{75} at correct tool selection and \pct{72} after argument checks. Qwen2.5-1.5B greedy exhibits a different pattern: all 75 recoverable objects are fully correct, but 25 responses cannot be recovered. Its sampling condition raises recovered validity to \pct{83} and finishes at \pct{79}. In contrast, recovered validity is at most \pct{9} for Pythia, \pct{8} for TinyLlama, and \pct{1} for Phi.

\begin{table*}[t]
\centering
\caption{Percentage passing each cumulative evaluation stage. Values are computed over all 100 prompts in each condition.}
\label{tab:components}
\small
\begin{tabular}{@{}llrrrrrr@{}}
\toprule
Model & Strategy & Recovered JSON & Correct tool & All args & Values match & No extras & Success \\
\midrule
Pythia-1.4B & greedy & 0 & 0 & 0 & 0 & 0 & 0 \\
Pythia-1.4B & sampling & 9 & 4 & 3 & 3 & 2 & 3 \\
Qwen2.5-0.5B & greedy & 82 & 75 & 72 & 72 & 72 & 72 \\
Qwen2.5-0.5B & sampling & 54 & 42 & 35 & 32 & 31 & 32 \\
Qwen2.5-1.5B & greedy & 75 & 75 & 75 & 75 & 75 & 75 \\
Qwen2.5-1.5B & sampling & 83 & 82 & 79 & 79 & 79 & 79 \\
TinyLlama-1.1B-Chat & greedy & 0 & 0 & 0 & 0 & 0 & 0 \\
TinyLlama-1.1B-Chat & sampling & 8 & 7 & 7 & 7 & 7 & 7 \\
Phi-1.5 & greedy & 0 & 0 & 0 & 0 & 0 & 0 \\
Phi-1.5 & sampling & 1 & 0 & 0 & 0 & 0 & 0 \\
\bottomrule
\end{tabular}
\end{table*}

The strict audit changes the operational interpretation. Across 1,000 generations, only five complete responses all from Qwen2.5-1.5B sampling are directly parseable as JSON. Therefore, even the best application-level rate relies on a recovery component that discards surrounding text or Markdown. A human may understand a fenced or prefaced object, but a strict consumer would reject it. This gap should be treated as part of the agent system, measured separately, and tested against adversarial or ambiguous brace content.

\subsection{Results by tool}

Table~\ref{tab:bytool} reports success for every tool category. Qwen2.5-1.5B greedy is perfect on calculation, task creation, and email, but scores \pct{0} on weather. Manual inspection of the saved outputs attributes these 20 weather failures to unrecoverable formatting rather than a general inability to copy the values. Sampling improves weather to \pct{70} while preserving strong results elsewhere. Qwen2.5-0.5B greedy is strongest on task creation (\pct{100}) and email (\pct{95}) and weakest on weather (\pct{50}). The uneven profile shows that an aggregate rate can conceal deterministic category failures.

\begin{table*}[t]
\centering
\caption{Notebook success rate (\%) by tool; each cell contains 20 prompts.}
\label{tab:bytool}
\small
\begin{tabular}{@{}llrrrrr@{}}
\toprule
Model & Strategy & Calculate & To-do & Weather & Search & Email \\
\midrule
Pythia-1.4B & greedy & 0 & 0 & 0 & 0 & 0 \\
Pythia-1.4B & sampling & 0 & 5 & 0 & 5 & 5 \\
Qwen2.5-0.5B & greedy & 55 & 100 & 50 & 60 & 95 \\
Qwen2.5-0.5B & sampling & 35 & 35 & 45 & 20 & 25 \\
Qwen2.5-1.5B & greedy & 100 & 100 & 0 & 75 & 100 \\
Qwen2.5-1.5B & sampling & 70 & 95 & 70 & 75 & 85 \\
TinyLlama-1.1B-Chat & greedy & 0 & 0 & 0 & 0 & 0 \\
TinyLlama-1.1B-Chat & sampling & 15 & 0 & 5 & 15 & 0 \\
Phi-1.5 & greedy & 0 & 0 & 0 & 0 & 0 \\
Phi-1.5 & sampling & 0 & 0 & 0 & 0 & 0 \\
\bottomrule
\end{tabular}
\end{table*}

\subsection{Failure modes}

\begin{figure*}[t]
  \centering
  \includegraphics[width=0.90\textwidth]{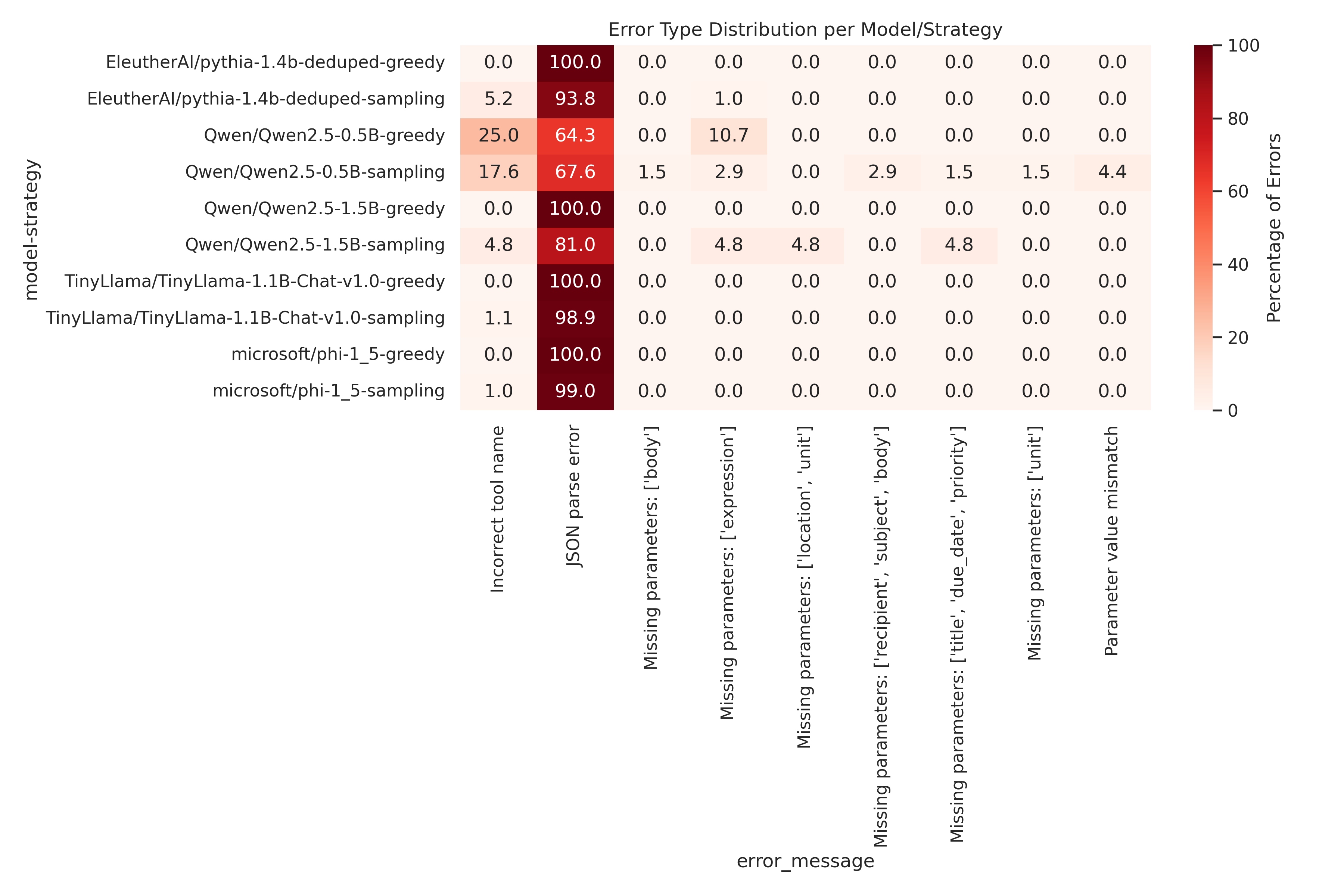}
  \caption{Distribution of terminal error labels among failed calls. The denominator differs by row because successful calls are excluded.}
  \label{fig:errors}
\end{figure*}

There are 658 failed notebook evaluations. Of these, 588 (\pct{89.4}) terminate at JSON recovery, 27 (\pct{4.1}) at the tool name, 40 (\pct{6.1}) because parameters are missing, and 3 (\pct{0.5}) because values mismatch. Percentages in Figure~\ref{fig:errors} are normalised within the failures of each model strategy condition; they should not be interpreted as percentages of all prompts. For example, every failure of Qwen2.5-1.5B greedy is a recovery failure, even though this represents 25 rather than 100 prompts.

Observed malformed outputs include Python examples instead of JSON, parameter arrays where objects are expected, surrounding explanations, repeated objects, and unfinished generations. The sequential evaluator assigns only the earliest detected error, so Figure~\ref{fig:errors} is not a multi-label taxonomy; later semantic faults may be hidden behind a parse or tool-name failure.

\subsection{Exploratory LangGraph diagnostic}

Table~\ref{tab:langgraph} presents the reduced LangGraph results using only fields needed to interpret the run. All ten attempts terminated with \texttt{Agent error: 'id'} before a model response or tool invocation was produced. Consequently, the \pct{0} success rates describe a framework-integration failure, not comparative model capability. The elapsed values measure time to failure in the wrapper and must not be compared with the CPU generation latencies in Table~\ref{tab:resources}. The larger Phi mean is primarily caused by its first attempt taking 34.703 ms; the remaining Phi attempts took approximately 1.4--1.8 ms.

\begin{table}[t]
\centering
\caption{Exploratory LangGraph integration diagnostic. Mean elapsed time is failure-handling time, not model inference latency.}
\label{tab:langgraph}
\small
\begin{tabular}{@{}lrrrr@{}}
\toprule
Model & Attempts & Success & Tool calls & Mean (ms) \\
\midrule
Phi-1.5 & 5 & \pct{0} & 0 & 8.222 \\
Pythia-1.4B & 5 & \pct{0} & 0 & 1.828 \\
\bottomrule
\end{tabular}
\end{table}

\subsection{Resource measurements}

\begin{figure*}[t]
  \centering
  \includegraphics[width=0.88\textwidth]{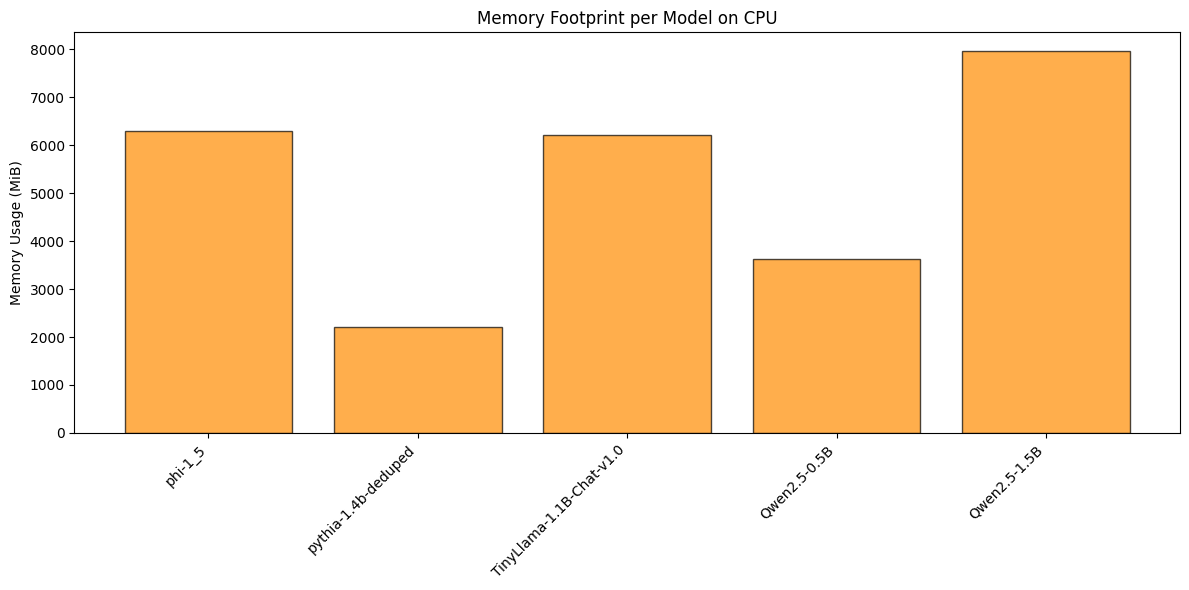}
  \caption{Process-level memory footprint recorded after loading each unquantised FP32 checkpoint on CPU. Because measurement occurred in a shared notebook process, the bars represent observed process RSS rather than isolated model weight size.}
  \label{fig:edge-memory}
\end{figure*}

\begin{figure*}[t]
  \centering
  \includegraphics[width=0.88\textwidth]{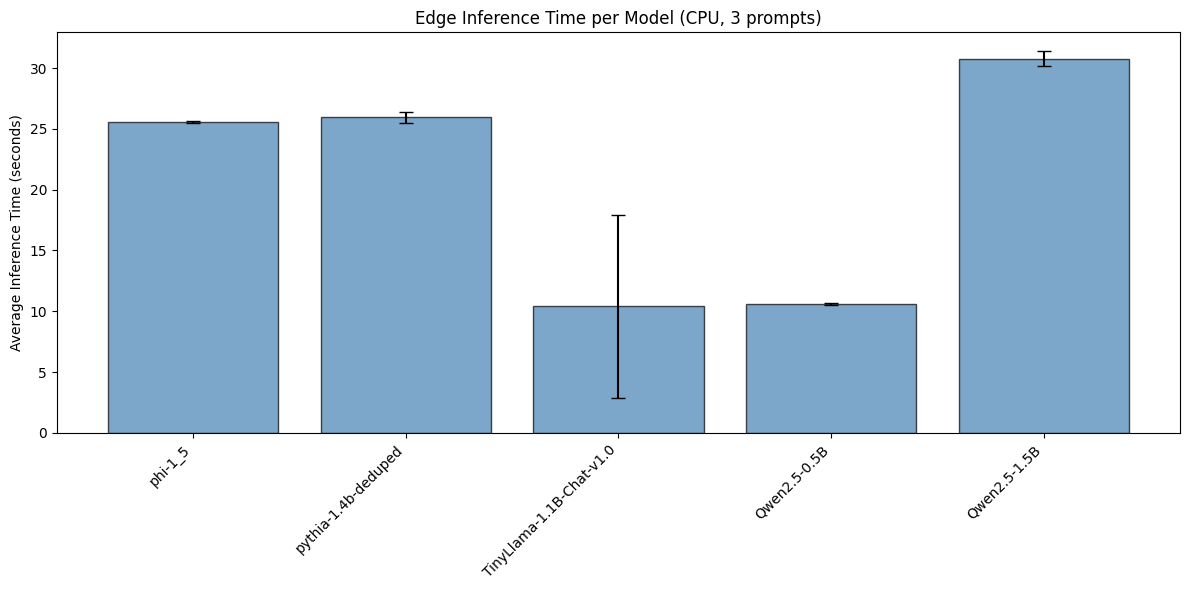}
  \caption{Mean CPU inference time for each model over the three resource-probe prompts. Error bars show the population standard deviation. Each greedy generation was capped at 50 new tokens after one warm-up generation.}
  \label{fig:edge-latency}
\end{figure*}

Table~\ref{tab:resources} reports the complete resource probe. Qwen2.5-1.5B has the highest observed RSS and latency, while Qwen2.5-0.5B has the shortest loading time and nearly the lowest inference time. The measurements do not scale monotonically with parameters: Pythia records the lowest RSS despite approximately 1.4B FP32 parameters, while Phi and TinyLlama record more than 6 GiB. This implausible ordering for isolated model storage reinforces that process RSS here includes allocator state, cached objects, and notebook history and should not be read as model weight size.
Figures~\ref{fig:edge-memory} and~\ref{fig:edge-latency} visualise the measured process memory and inference latency, respectively.

\begin{table*}[t]
\centering
\caption{Separate CPU resource probe. Latency is mean $\pm$ population standard deviation over three prompts after one warm-up; generation was capped at 50 new tokens.}
\label{tab:resources}
\small
\begin{tabular}{@{}lrrrr@{}}
\toprule
Model & Process RSS (MiB) & Load time (s) & Inference (s) & Quantised \\
\midrule
Phi-1.5 & 6,306.98 & 57.59 & $25.539 \pm 0.091$ & No \\
Pythia-1.4B & 2,201.38 & 195.88 & $25.950 \pm 0.455$ & No \\
TinyLlama-1.1B-Chat & 6,222.11 & 118.76 & $10.415 \pm 7.526$ & No \\
Qwen2.5-0.5B & 3,637.09 & 54.31 & $10.627 \pm 0.080$ & No \\
Qwen2.5-1.5B & 7,960.08 & 218.11 & $30.782 \pm 0.588$ & No \\
\bottomrule
\end{tabular}
\end{table*}

Within the measured setup, Qwen2.5-0.5B greedy offers the strongest practical compromise: its \pct{72} recovered-call rate is three points below Qwen2.5-1.5B greedy, while its measured mean latency is roughly \pct{65} lower and RSS roughly \pct{54} lower. This is an engineering comparison within one notebook run, not a claim about other CPUs, quantisation formats, runtimes, or output lengths.

\section{Discussion}

\subsection{What determines reliability?}

Parameter count alone does not explain the results. Models near 1.4B parameters range from \pct{0} to \pct{79}, while the 0.5B Qwen checkpoint reaches \pct{72} under greedy decoding. Training objective, instruction-following behaviour, tokenizer, model family, and prompt compatibility are confounded. The result therefore supports checkpoint-level selection and task-specific evaluation, not the claim that a particular parameter threshold guarantees agentic capability.

Decoding also interacts with the checkpoint. Sampling benefits Qwen2.5-1.5B slightly and creates a few successes for Pythia and TinyLlama, but damages Qwen2.5-0.5B by 40 percentage points. For deterministic automation, this variability argues for validating the exact model prompt decoder combination rather than adopting a universal temperature recommendation. Repeated sampling runs are needed before attributing small differences such as \pct{75} versus \pct{79} to a stable effect.

Most importantly, the recovery layer is part of the measured system. Reporting only recovered success would overstate raw instruction compliance; reporting only strict JSON would ignore a simple application mechanism that recovers useful calls. Both are operationally relevant. A robust deployment should use a formally specified parser, schema validation, bounded repair attempts, and rejection on ambiguity. Grammar- or schema-constrained decoding is a promising comparison condition, but syntactic guarantees do not establish semantic correctness \citep{geng2025jsonschema}.

\subsection{Deployment implications}

\textbf{Why an edge agent should not rely on an SLM alone.} An SLM should be treated as a fallible proposal generator rather than the sole decision and execution authority of an edge agent. Even in this favourable benchmark, where every prompt explicitly supplied the correct tool and all required values, the best recovered-call rate was only \pct{79}, and strict whole-response JSON validity was only \pct{5} in that best condition. Real deployments introduce ambiguous requests, missing information, distractor tools, changing state, sensor noise, adversarial inputs, and network or tool failures, so the effective error rate can be expected to increase. Furthermore, syntactically valid JSON does not guarantee that the selected action is safe, authorised, or contextually correct, while sampling makes behaviour non-deterministic and limited edge memory and compute can restrict redundancy, verification, and timely recovery. A single mistaken email recipient, actuator command, file operation, or API argument may create consequences disproportionate to the convenience of local inference. Evidence from function-calling robustness and agent-safety research similarly shows sensitivity to query variation, parameter errors, and harmful tool interactions \citep{wang2025hammerbench,rabinovich2025robustness,ruan2023toolemu}. Accordingly, the SLM should propose a call to a deterministic supervisory layer that enforces schemas, permissions, state constraints, confidence thresholds, and human confirmation before any consequential action is executed.

The benchmark contains mock email and task-creation actions, which illustrate why parseability is only a first gate. Before execution, an agent should verify that the selected tool is allow-listed, arguments satisfy types and ranges, recipients and resources fall within policy, and the user has authority to perform the action. Consequential or irreversible operations should require confirmation. Tools should expose least privilege, idempotency where possible, dry-run modes, timeouts, and auditable logs. Failure should default to no action rather than best-effort interpretation.

Qwen2.5-1.5B's recovered success is encouraging for a bounded local assistant, but a \pct{21} failure rate under its best observed condition is unacceptable for unmonitored actions. The strict audit makes the case stronger: without recovery, \pct{95} of even that condition is not directly machine-readable. Agent safety research likewise finds that realistic tool access creates privacy and financial risks that are not captured by benign accuracy tests \citep{ruan2023toolemu}. Reliability controls must therefore sit outside the generative model.

\section{Limitations and Threats to Validity}

\textbf{Construct validity.} Prompts explicitly reveal the correct tool and arguments, so ``tool selection'' largely measures copying and serialization. Calls are not executed, schema types and enums are not formally validated, and the notebook success metric permits extra parameter keys. The parser's greedy brace extraction can also merge multiple objects or accept surrounding prose. No refusal, clarification, irrelevant-tool, ambiguous-intent, adversarial, or safety cases are included.

\textbf{Internal validity.} Checkpoints differ in training and alignment; some are base models and one is explicitly chat-tuned. Chat templates were not applied. Sampling has only one run per prompt. Model commit hashes and exact package versions were not recorded, limiting exact replication. Generation length is capped, and truncation may be counted as a syntax failure.

\textbf{Statistical validity.} The 100 prompts are manually templated from 20 value combinations per tool rather than sampled from a documented population. Wilson intervals quantify binomial uncertainty conditional on this item set but do not represent the diversity of real user requests. Tool-level cells contain only 20 observations, and reported McNemar tests are exploratory.

\textbf{Systems validity.} The resource probe uses only three prompts, a shared notebook process, and unspecified CPU hardware. RSS includes more than model weights, load time may include cache and I/O effects, and fixed token caps do not control actual generated length. No quantised runtime, mobile device, energy, throughput, or memory-bandwidth measurement is included. The results should not be generalised to edge hardware without a controlled rerun.

\textbf{External validity.} This is not an MCP protocol benchmark or a realistic agent trajectory. It contains five tools, one call per prompt, no distractors, no tool descriptions in context, no intermediate results, and no live failures. HammerBench, MCPToolBench++, and function-call robustness benchmarks indicate the broader scenarios needed for external validation \citep{wang2025hammerbench,fan2025mcptoolbench,rabinovich2025robustness}.

\FloatBarrier
\section{Future Work}

The next evaluation should (1) record hardware, software, revisions, token counts, and energy; (2) run each stochastic condition across multiple seeds; (3) compare base, instruction-tuned, and function-calling-tuned checkpoints under their official chat templates; (4) add strict JSON Schema validation and constrained decoding; (5) include distractor and irrelevant tools, paraphrases, missing arguments, and clarification cases; (6) execute safe tools in a sandbox and score deterministic state changes; and (7) implement MCP discovery, tool outputs, failures, context-length variation, and multi-step trajectories. Quantised runtimes should then be benchmarked in isolated processes on representative edge platforms, including Raspberry Pi, NVIDIA Jetson Nano, Arduino UNO Q, Orange Pi, and LattePanda, with peak and steady-state memory, energy, and thermal behaviour measured separately.

\section{Conclusion}

This controlled study finds that sub-2B models vary dramatically in JSON-based tool invocation. Qwen2.5 checkpoints substantially outperform similarly sized Phi, Pythia, and TinyLlama checkpoints, and decoding strategy has effects ranging from strongly harmful to mildly beneficial. The strongest notebook result is Qwen2.5-1.5B sampling at \pct{79}, but it carries the largest measured resource cost and depends on response recovery: only five of all 1,000 raw generations are strict whole-response JSON. The evidence supports SLMs as components in bounded local tool systems, not as trusted autonomous executors. Safe deployment requires constrained interfaces, schema and policy validation, least-privilege tools, explicit confirmation for consequential actions, and human escalation when confidence or input completeness is insufficient.

\section*{Acknowledgements}

The authors acknowledge the open-source model and software communities whose resources enabled this study. The experimental notebook and released benchmark artifacts are publicly available in the project repository at \url{https://github.com/Abrar051/sml_mcp_benchmark}.

\section*{Reproducibility and Data Availability}

The experimental artifact supplied with this study contains the notebook, 100 benchmark prompts, 1,000 detailed model responses, summary metrics, error counts, and figures. The files are \texttt{mcp\_test\_final.ipynb}, \texttt{benchmark\_prompts.csv}, \texttt{detailed\_results.csv}, \texttt{metrics\_summary.csv}, \texttt{error\_counts.csv}, \texttt{success\_rates.png}, \texttt{error\_heatmap.png}, \texttt{image.png}, \texttt{image\_2.png}, and \texttt{flowImage.png}. The public project repository is available at \url{https://github.com/Abrar051/sml_mcp_benchmark}. A versioned archival release and persistent identifier should be created before final submission to support long-term reproducibility. No human participants or personal datasets were used; email addresses and actions are synthetic.


\begin{thebibliography}{99}

\bibitem[Arduino(n.d.)]{arduino2026unoq}
Arduino (n.d.) `Debian Linux basics for UNO Q'. Available at: \url{https://docs.arduino.cc/tutorials/uno-q/debian-guide} (Accessed: 2 September 2026).

\bibitem[Biderman et~al.(2023)]{biderman2023pythia}
Biderman, S., Schoelkopf, H., Anthony, Q., Bradley, H., O'Brien, K., Hallahan, E., Khan, M.A., Purohit, S., Prashanth, U.S.V.S.N.S., Raff, E., Skowron, A., Sutawika, L. and van der Wal, O. (2023) `Pythia: A suite for analyzing large language models across training and scaling', \emph{arXiv preprint}, arXiv:2304.01373. Available at: \url{https://arxiv.org/abs/2304.01373} (Accessed: 2 September 2026).

\bibitem[Chen et~al.(2025)]{chen2025elib}
Chen, H., Tian, C., He, Z., Yu, B., Liu, Y. and Cao, J. (2025) `Inference performance evaluation for LLMs on edge devices with a novel benchmarking framework and metric', \emph{arXiv preprint}, arXiv:2508.11269. Available at: \url{https://arxiv.org/abs/2508.11269} (Accessed: 2 September 2026).

\bibitem[Chen et~al.(2025)]{chen2025octopus}
Chen, W., Li, Z. and Ma, M. (2025) `Octopus: On-device language model for function calling of software APIs', in \emph{Proceedings of the 2025 Conference of the Nations of the Americas Chapter of the Association for Computational Linguistics: Human Language Technologies, Industry Track}, pp. 329--339. Association for Computational Linguistics. Available at: \url{https://aclanthology.org/2025.naacl-industry.27/} (Accessed: 2 September 2026).

\bibitem[Erdogan et~al.(2024)]{erdogan2024tinyagent}
Erdogan, L.E., Lee, N., Jha, S., Kim, S., Tabrizi, R., Moon, S., Hooper, C.R.C., Anumanchipalli, G., Keutzer, K. and Gholami, A. (2024) `TinyAgent: Function calling at the edge', in \emph{Proceedings of the 2024 Conference on Empirical Methods in Natural Language Processing: System Demonstrations}, pp. 80--88. Association for Computational Linguistics. Available at: \url{https://aclanthology.org/2024.emnlp-demo.9/} (Accessed: 2 September 2026).

\bibitem[Fan et~al.(2025)]{fan2025mcptoolbench}
Fan, S., Ding, X., Zhang, L. and Mo, L. (2025) `MCPToolBench++: A large scale AI agent Model Context Protocol MCP tool use benchmark', \emph{arXiv preprint}, arXiv:2508.07575. Available at: \url{https://arxiv.org/abs/2508.07575} (Accessed: 2 September 2026).

\bibitem[Geng et~al.(2025)]{geng2025jsonschema}
Geng, S., Cooper, H., Moskal, M., Jenkins, S., Berman, J., Ranchin, N., West, R., Horvitz, E. and Nori, H. (2025) `Generating structured outputs from language models: Benchmark and studies', \emph{arXiv preprint}, arXiv:2501.10868. Available at: \url{https://arxiv.org/abs/2501.10868} (Accessed: 2 September 2026).

\bibitem[Jhandi et~al.(2025)]{jhandi2025toolcalling}
Jhandi, P., Kazi, O., Subramanian, S. and Sendas, N. (2025) `Small language models for efficient agentic tool calling: Outperforming large models with targeted fine-tuning', \emph{arXiv preprint}, arXiv:2512.15943. Available at: \url{https://arxiv.org/abs/2512.15943} (Accessed: 2 September 2026).

\bibitem[Kandala et~al.(2024)]{kandala2024tinyllm}
Kandala, S.V., Medaranga, P. and Varshney, A. (2024) `TinyLLM: A framework for training and deploying language models at the edge computers', \emph{arXiv preprint}, arXiv:2412.15304. Available at: \url{https://arxiv.org/abs/2412.15304} (Accessed: 2 September 2026).

\bibitem[LattePanda(n.d.)]{lattepanda2026}
LattePanda (n.d.) `LattePanda documentation: Overview'. Available at: \url{https://docs.lattepanda.com/} (Accessed: 2 September 2026).

\bibitem[Li et~al.(2023)]{li2023apibank}
Li, M., Zhao, Y., Yu, B., Song, F., Li, H., Yu, H., Li, Z., Huang, F. and Li, Y. (2023) `API-Bank: A comprehensive benchmark for tool-augmented LLMs', in \emph{Proceedings of the 2023 Conference on Empirical Methods in Natural Language Processing}, pp. 3102--3116. Association for Computational Linguistics. Available at: \url{https://aclanthology.org/2023.emnlp-main.187/} (Accessed: 2 September 2026).

\bibitem[Li et~al.(2023)]{li2023phi}
Li, Y., Bubeck, S., Eldan, R., Del Giorno, A., Gunasekar, S. and Lee, Y.T. (2023) `Textbooks are all you need II: phi-1.5 technical report', \emph{arXiv preprint}, arXiv:2309.05463. Available at: \url{https://arxiv.org/abs/2309.05463} (Accessed: 2 September 2026).

\bibitem[Lin et~al.(2025)]{lin2024hammer}
Lin, Q., Wen, M., Peng, Q., Nie, G., Liao, J., Wang, J., Mo, X., Zhou, J., Cheng, C., Zhao, Y., Wang, J. and Zhang, W. (2025) `Hammer: Robust function-calling for on-device language models via function masking', in \emph{Proceedings of the Thirteenth International Conference on Learning Representations}. Available at: \url{https://openreview.net/forum?id=yVQcr4qjD6} (Accessed: 2 September 2026).

\bibitem[Liu et~al.(2024)]{liu2024mobilellm}
Liu, Z., Zhao, C., Iandola, F., Lai, C., Tian, Y., Fedorov, I., Xiong, Y., Chang, E., Shi, Y., Krishnamoorthi, R., Lai, L. and Chandra, V. (2024) `MobileLLM: Optimizing sub-billion parameter language models for on-device use cases', in \emph{Proceedings of the 41st International Conference on Machine Learning}, \emph{Proceedings of Machine Learning Research}, 235, pp. 32431--32454. Available at: \url{https://proceedings.mlr.press/v235/liu24ce.html} (Accessed: 2 September 2026).

\bibitem[Lu et~al.(2025)]{lu2025slm}
Lu, Z., Li, X., Cai, D., Yi, R., Liu, F., Zhang, X., Lane, N.D. and Xu, M. (2025) `Small language models: Survey, measurements, and insights', \emph{arXiv preprint}, arXiv:2409.15790. Available at: \url{https://arxiv.org/abs/2409.15790} (Accessed: 2 September 2026).

\bibitem[Lu et~al.(2025)]{lu2025toolsandbox}
Lu, J., Holleis, T., Zhang, Y., Aumayer, B., Nan, F., Bai, H., Ma, S., Ma, S., Li, M., Yin, G., Wang, Z. and Pang, R. (2025) `ToolSandbox: A stateful, conversational, interactive evaluation benchmark for LLM tool use capabilities', in \emph{Findings of the Association for Computational Linguistics: NAACL 2025}, pp. 1160--1183. Association for Computational Linguistics. Available at: \url{https://aclanthology.org/2025.findings-naacl.65/} (Accessed: 2 September 2026).

\bibitem[NVIDIA(n.d.)]{nvidia2026jetson}
NVIDIA (n.d.) `Jetson Nano'. Available at: \url{https://developer.nvidia.com/embedded/jetson-nano} (Accessed: 2 September 2026).

\bibitem[Orange Pi(n.d.)]{orangepi2026}
Orange Pi (n.d.) `Orange Pi 5'. Available at: \url{https://www.orangepi.org/html/hardWare/computerAndMicrocontrollers/details/Orange-Pi-5.html} (Accessed: 2 September 2026).

\bibitem[Patil et~al.(2023)]{patil2023gorilla}
Patil, S.G., Zhang, T., Wang, X. and Gonzalez, J.E. (2023) `Gorilla: Large language model connected with massive APIs', \emph{arXiv preprint}, arXiv:2305.15334. Available at: \url{https://arxiv.org/abs/2305.15334} (Accessed: 2 September 2026).

\bibitem[Qwen Team(2024)]{qwen2024qwen25}
Qwen Team (2024) `Qwen2.5 technical report', \emph{arXiv preprint}, arXiv:2412.15115. Available at: \url{https://arxiv.org/abs/2412.15115} (Accessed: 2 September 2026).

\bibitem[Raspberry Pi Ltd(2026)]{raspberrypi2026}
Raspberry Pi Ltd (2026) `Raspberry Pi 5 product brief'. Available at: \url{https://pip.raspberrypi.com/documents/RP-008348-DS-raspberry-pi-5-product-brief.pdf} (Accessed: 2 September 2026).

\bibitem[Rabinovich and Anaby-Tavor(2025)]{rabinovich2025robustness}
Rabinovich, E. and Anaby-Tavor, A. (2025) `On the robustness of agentic function calling', \emph{arXiv preprint}, arXiv:2504.00914. Available at: \url{https://arxiv.org/abs/2504.00914} (Accessed: 2 September 2026).

\bibitem[Ruan et~al.(2024)]{ruan2023toolemu}
Ruan, Y., Dong, H., Wang, A., Pitis, S., Zhou, Y., Ba, J., Dubois, Y., Maddison, C.J. and Hashimoto, T. (2024) `Identifying the risks of LM agents with an LM-emulated sandbox', in \emph{Proceedings of the Twelfth International Conference on Learning Representations}. Available at: \url{https://openreview.net/forum?id=GEcwtMk1uA} (Accessed: 2 September 2026).

\bibitem[Schick et~al.(2023)]{schick2023toolformer}
Schick, T., Dwivedi-Yu, J., Dess\`i, R., Raileanu, R., Lomeli, M., Hambro, E., Zettlemoyer, L., Cancedda, N. and Scialom, T. (2023) `Toolformer: Language models can teach themselves to use tools', \emph{Advances in Neural Information Processing Systems}, 36, pp. 68539--68551. Available at: \url{https://proceedings.neurips.cc/paper_files/paper/2023/hash/d842425e4bf79ba039352da0f658a906-Abstract-Conference.html} (Accessed: 2 September 2026).

\bibitem[Wang et~al.(2025)]{wang2025hammerbench}
Wang, J., Zhou, J., Wen, M., Mo, X., Zhang, H., Lin, Q., Jin, C., Wang, X., Zhang, W., Peng, Q. and Wang, J. (2025) `HammerBench: Fine-grained function-calling evaluation in real mobile device scenarios', \emph{arXiv preprint}, arXiv:2412.16516. Available at: \url{https://arxiv.org/abs/2412.16516} (Accessed: 2 September 2026).

\bibitem[Yao et~al.(2025)]{yao2025taubench}
Yao, S., Shinn, N., Razavi, P. and Narasimhan, K.R. (2025) `$\tau$-bench: A benchmark for tool--agent--user interaction in real-world domains', in \emph{Proceedings of the Thirteenth International Conference on Learning Representations}. Available at: \url{https://openreview.net/forum?id=roNSXZpUDN} (Accessed: 2 September 2026).

\bibitem[Zhang et~al.(2024)]{zhang2024tinyllama}
Zhang, P., Zeng, G., Wang, T. and Lu, W. (2024) `TinyLlama: An open-source small language model', \emph{arXiv preprint}, arXiv:2401.02385. Available at: \url{https://arxiv.org/abs/2401.02385} (Accessed: 2 September 2026).

\end{thebibliography}
\end{document}